# Segmentation and Analysis of a Sketched Truss Frame Using Morphological Image Processing Techniques

**MirSalar Kamari[1], Oğuz Güneş[2]**

1. Masters' Student, Department of Civil Engineering, Istanbul Technical University, kamari@itu.edu.tr
2. Assistant Professor, Department of Civil Engineering, Istanbul Technical University, ogunes@itu.edu.tr

**Abstract**

Development of computational tools to analyze and assess the building capacities has had a major impact in civil engineering. The interaction with the structural software packages is becoming easier and the modeling tools are becoming smarter by automating the users' role during their interaction with the software. One of the difficulties and the most time-consuming steps involved in the structural modeling is defining the geometry of the structure to provide the analysis. This paper is dedicated to the development of a methodology to automate analysis of a hand-sketched or computer generated truss frame drawn on a piece of paper. First, we focus on the segmentation methodologies for hand sketched truss components using the morphological image processing techniques, and then we provide a real time analysis of the truss. We visualize and augment the results on the input image to facilitate the public understanding of the truss geometry and internal forces. MATLAB is used as the programming language for the image processing purposes, and the truss is analyzed using Sap2000 API to integrate with MATLAB to provide a convenient structural analysis. This paper highlights the potential of the automation of the structural analysis using image processing to quickly assess the efficiency of structural systems. Further development of this frame work is likely to revolutionize the way that structures are modeled and analyzed.

# Introduction

The need for the innovative tools to leverage the task automation has been beneficial to human beings from the ancient time. From invention of typing machine to arrival of the autonomous vehicles, human beings have been always appreciating the automation and the self-governing systems. With the advent of the modern computer to handle the massive calculations, most of the tasks in one's life are easier than ever before. We are motived by the need for the algorithms that can autonomously understand and label the structural details from the building images and pass the information to structural solvers and software to quickly assess the structural performance of structures. This can revolutionize the way that the buildings are analyzed.

With the advent of the Application Programming Interface (API), codes and algorithms can now be integrated with the currently available software to enhance the workability of them through automating links in their overall processes. In this paper, the image of a hand-sketched or computer generated truss containing its structural information is given to an algorithm to read the characters and segment the truss members and components. Then, the segmentation results are sent to Sap2000 (a structural solver) using an API to carry out the analysis and retrieve back the analysis results. Then, the analysis results, for instance internal forces, are overlaid ~~of~~ on the input image.

Trusses are structural elements built from assemblage of steel members connected to each other. Since the connection of these members is not rigid and cannot carry bending moments, the only unknown in any truss element is the axial load. Usually most of the documented truss problems are laid in two dimensions which are referred to as the planer trusses. We assume the truss problem to be a planer truss. A typical truss contains members, joint elements and the supports. To draw the geometry of a truss in the structural analysis books, particular notations are used to represent the different components of it. Mostly, the supports are shown with triangles. To specify the direction of the degree of freedom in any support a line is drawn at the side of the triangle parallel to its lateral. In most of the structural analysis books, joints are shown in filled circles to connect the members. Also, loads on the trusses, which are applied to the joints, are represented with arrows to show its applied direction. Figure 1 shows the common notations used to schematically draw the trusses. These morphological component properties are studied to segment and recognize the geometry of the truss through strategies mentioned in the following sections.

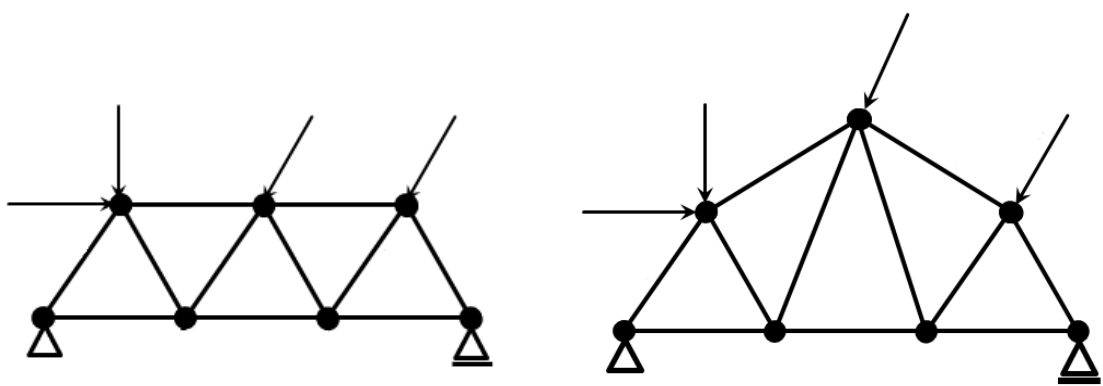

**Figure 1. Typical notations used to schematically draw trusses**

## The methodology

The image processing toolbox of MATLAB is used for image processing in this study. The morphological properties of the input image are studied to segment and recognize the geometrical information of the drawn truss object. The segmentation criterion for each component is explained at the relevant section. To provide the structural analysis for the detected truss, Sap2000 is called form MATLAB to get the geometry and loading information as well as the boundary conditions to retrieve the analysis results and the internal forces in back in MATLAB. Figure2 shows the general order of segmentation and framework proposed in this research.

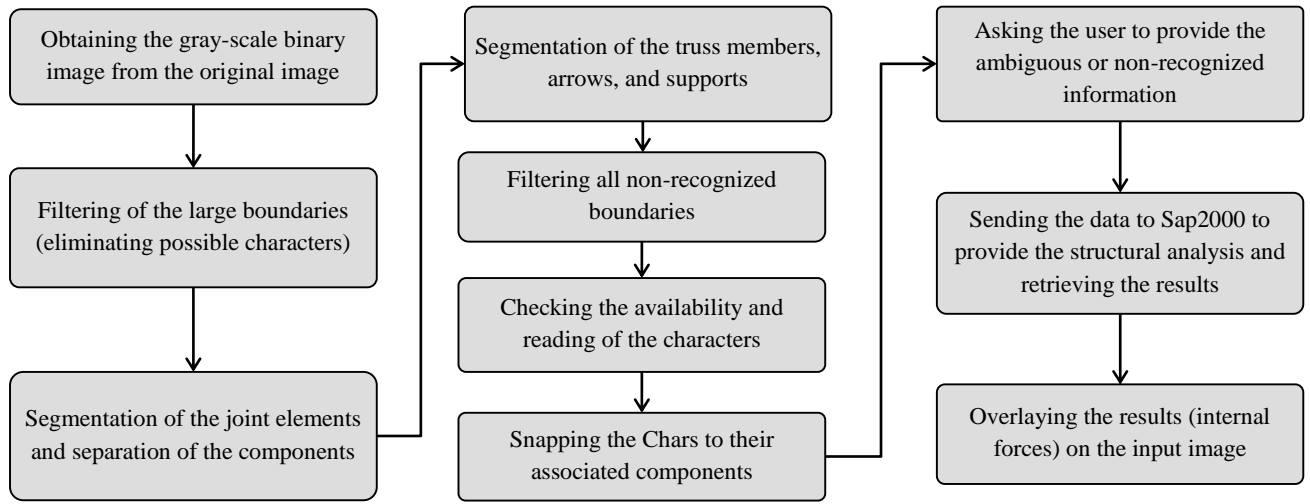

**Figure 2. The general outline of the algorithm**

The essential and key steps of the methodology are as follows (Figure 2):
1- Generating the binary gray-scale image and eliminating the text characters
2- Segmentation of the truss components
3- Detection and recognition of the character strings and designation of them to the proper components
4- Confirmation for the data validation
5- Conduction of the structural analysis using Sap2000 API
6- Overlaying the results on the input image

These six steps are explained below in explicit details.

## 1.Generating the binary gray-scale image and eliminating the text characters

The input image (Figure 3a) is converted to a gray-scale binary image (Figure 3b). To increase the accuracy of the detection of the truss components, characters and generally the boundaries with lower areas are removed from the boundary image (Figure 3c). They will be added on the binary image in the following steps to be processed and recognized.

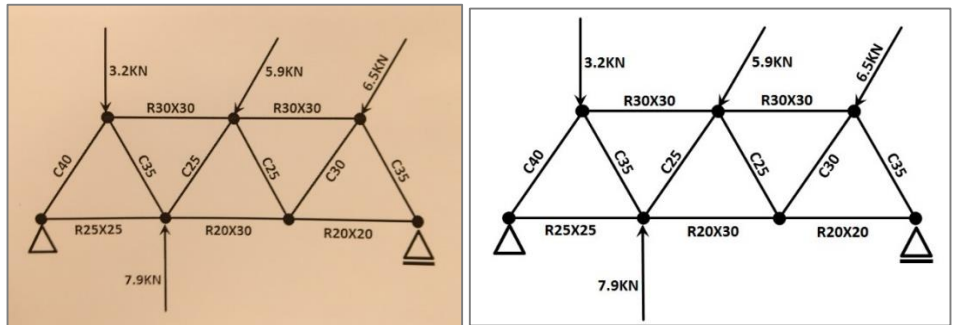

(a). The original image     (b) The gray-scale binary image

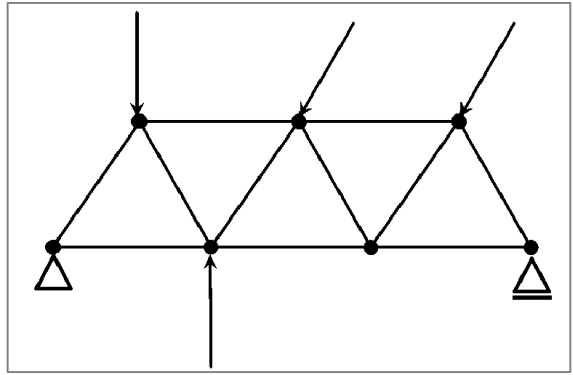

(c). The lower boundary areas (texts characters) are removed
Figure 3: Generating the binary gray-scale image and eliminating the text characters

## 2. Segmentation of the truss components

To model the geometry of the drawn truss, its components will have to be segmented to identify their location in the image and understand how these components are connected to each other. The segmentation methodologies and strategies are explained in the following sections:

### 2.1. Segmentation of the joint elements

All the boundary edges around each object in the gray-scale boundary image are traced and detected to perform the erosion of the image. The erosion operation can be explained with a mathematical expression as equation (1):

$$A \ominus B = \{\ z\ |\ B_z \subseteq A\} \tag{1}$$

where, A is the input image, B is the structuring element, $B_z$ is the transferred structuring element, and, z is the sets of points that meet the $B_z \subseteq A$ criterion. In other words, the space z contains all the pixels that transferred $B_z$ pixels fits fully in the A space [1]. It is assumed that the joint elements are large in diameter to survive the erosion operation. The binary gray-scale input image is eroded (Figure 4) to detect and obtain coordinates of all the joint elements. After eroding the image all the remaining objects, with lower boundary area, are considered to be the joint elements. The locations and the coordinates of the all detected joint elements are stored to model the geometric shape of the truss in the following steps. Since the joint elements are the gap between the supports, members and loading arrows, the detection of the supports will assist to recognize the remaining components by separating them from each other making it easy to define the recognition function. The recognized joint elements are labeled and shown in Figure 4.

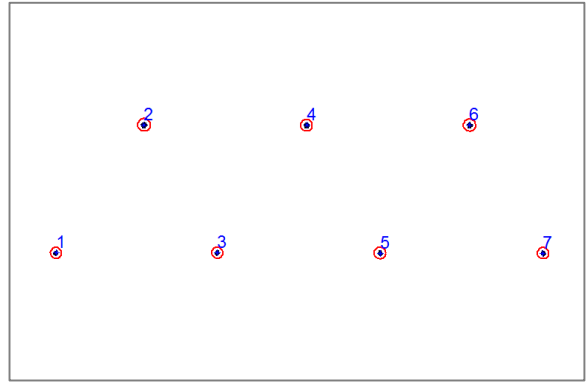

**Figure4. Detected joint objects following the erosion of the binary image**

## 2.2. Segmentation of the line elements (truss members)

After localization and segmentation of the joint elements the availability of the line elements is checked between two joint elements. The pixel values between the centers of the two joint elements are examined to confirm the availability of a line element. It is also assumed that the line elements are drawn in a straight fashion, from center to center of two joint elements; since any inclination in the drawing of the line element will misrepresent pixel sampling studies between two joint elements. If three points (say, point 1, 2, and 3) are straight along each other and two pairs of lines connect point 1 to 2 and point 2 to 3, the line connecting point 1 to 3 is neglected, even though the pixels between them meet the criteria for the availability of a line (Figure 5a). Next, the lines and the joint elements are masked and, subtracted from the binary image (Figure 5b) isolating the remaining components of the truss, increasing the accuracy of detection of the remaining components. The line elements are then segmented and labeled, as shown in Figure 5a.

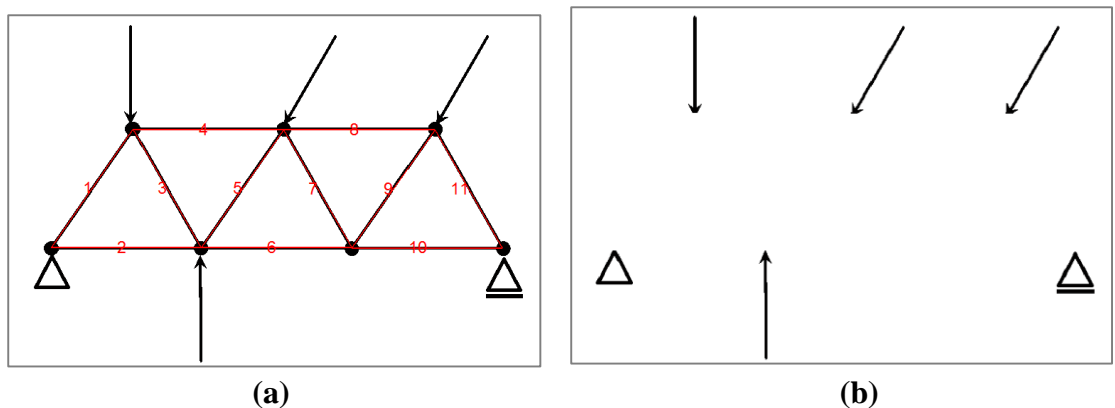

(a)          (b)

**Figure 5. (a) Detection, coloring and labeling of the line elements. (b) Subtraction of the segmented joints and line elements**

## 2.3. Segmentation of the arrows

After detection of the joint and the line elements and their elimination from the input image, arrows are detected in the binary image (Figure 5b). Morphological properties of the arrows can be explained as the line shaped objects with their centroid shifted to their either ends. The algorithm will study the remaining boundaries' centroids and also, it will assess the boundaries' similarity to a straight line. Thus the following two criteria will have to be met to consider a boundary as an arrow:

- The line similarity value of the object will have to be greater than 0.95. This criterion will select all the objects similar to a line object.
- The centroid shift value calculated from (2) will have to be greater than 0.01. This criterion will select all the boundaries with their centroid shifted from their bounding box center.

$$Centroid\ Shift = \frac{\sqrt{(Bx - Cx)^2 + (By - Cy)^2}}{\sqrt{(LBx^2 - LBy^2)}} \qquad (2)$$

where, $(Bx, By)$ are the coordinates of bounding box center, $(Cx, Cy)$ are the centroid coordinates of the boundary and $LBx, LBy$ are the length and width of bounding box. A hazard map (Figure 6a) containing lines, supports and arrows, is proposed to study the accuracy of the algorithm. Figure 6b shows the set, containing all the objects with a shape similar to a line object. Figure 6c contains all the objects that their centroids are shifted from the center of the smallest bounding box surrounding the object, calculated from equation (2). Figure 6d, is the intersection of the boundaries selected in Figures 6b and 6c and shows the result of the segmentation.

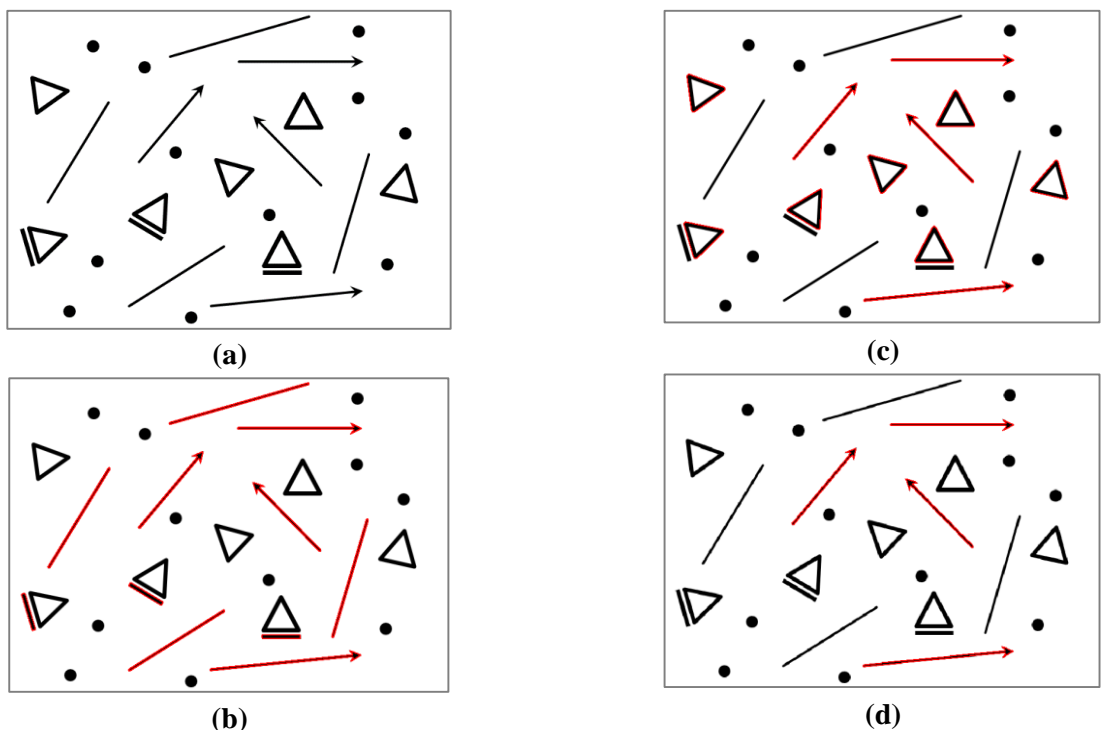

Figure 6. (a) The hazard map containing pinned and roller supports, lines, arrows and joint objects, (b) Set of all objects similar to line object. (c) Set of objects with their centroid shifted, (d) Arrow segmentation results, the intersection of the objects in (Figure 6b) and (Figure 6c)

The orientation of arrows is calculated from the position of their centroids. The closest bounding box corner to the arrow's centroid is determined for all the arrow boundaries, and then, the orientation angle is calculated from the orientation of the line that connects the center box to the closest bounding box corner. The joint element to which the arrow or the load is applied can be determined form the position of the tip of the arrow and the position of the joint element. The closest joint element to the tip arrow is considered the joint element to which the load is applied. To reduce the confusion rate of the algorithm after detection of the arrows all of them are subtracted from the binary image.

## 2.4. Segmentation of the supports

After segmentation and removal of the joints, lines, and arrows the segmentation of the support elements can be conducted by filling all the holes of the remaining boundaries in the gray-scale image. (Figure 7a)

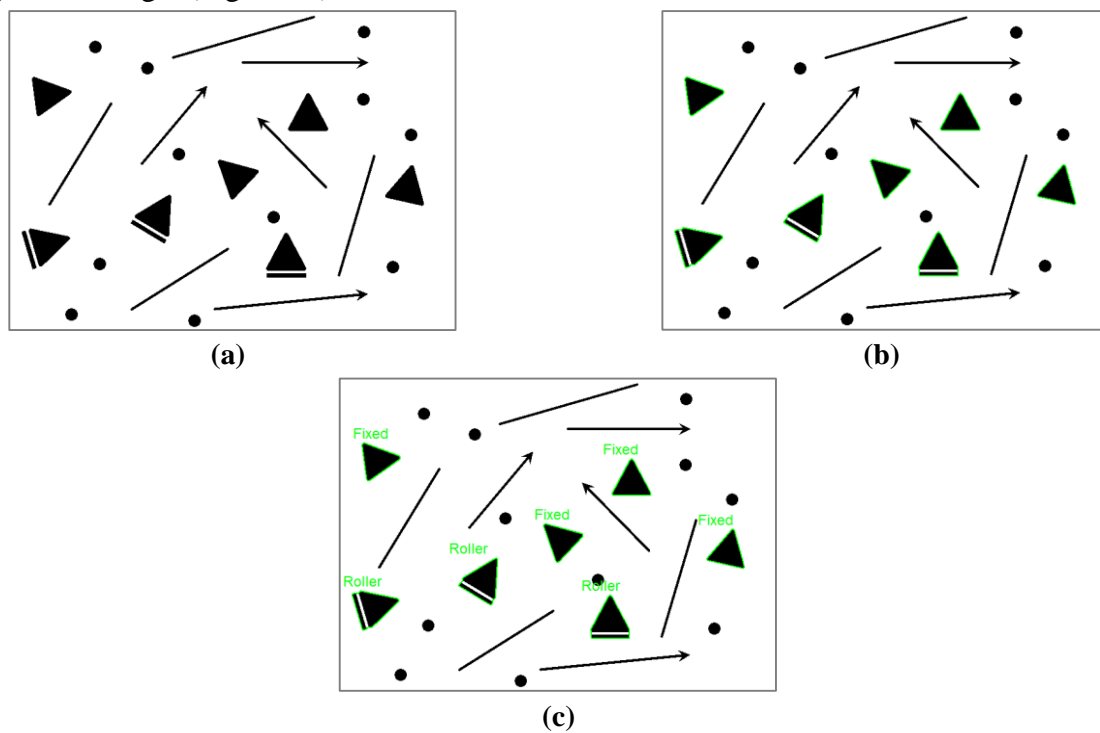

**Figure 7 (a) Gray-scaly binary hazard map with all its holes filled. (b) All the boundaries with their area over bounding box area that ranges from 0.65 to 0.75 are shown in the bounding box. (c) The result of the support segmentation.**

All the supports are assumed to have a triangular shape. The algorithm is capable of detecting the pinned and the roller supports. To specify the direction of the degree of freedom in the roller supports, a line parallel to the lateral side of the triangle is assumed to be drawn. First, to segment the supports, the algorithm detects all the triangles. Then it separates the pinned supports from the roller ones and associates the direction of the degree of freedom to the roller supports. The bounding box, centroid and the area of each boundary are studied to detect the support objects. For an object to be considered as a support the following criteria will have to be met:

- The ratio of the boundary area over area of the smallest bounding box surrounding the object has to range from 0.65 to 0.75. All the boundaries satisfying this criterion are shown in Figure 7b.
- The centroid location of the boundary will have to be shifted from the location of the smallest bounding box center which can be calculated form equation (1). This criterion will have to be met to avoid the selection of any object with circular shape that survived the previous filter. All the objects with shifted centroid form their bounding box centers are shown in Figure 6c.

The above criteria will segment the triangles from the remaining objects. The algorithm will seek a line around the triangles to segment pinned support from the roller. All the triangles are dilated with structuring circle element with diameter of 20 percent of the maximum length of the bounding box containing that triangle. The dilation can be written as a mathematical formula as:

$$A \oplus B = \{\; z \;|\; B_z \cap A \subseteq A\} \qquad (3)$$

where, A is the input binary boundary that will be dilated, B is the structuring element, $B_z$ is the transferred structuring element, and z is the sets of points that meet the $B_z \cap A \subseteq A$ criterion [1]. The dilation process will increase the boundaries of the triangles to collide with their surroundings. Possible roller candidates will collide with their lines to unite the line and triangle boundaries. The boundaries flagged with the line collision are the possible roller support boundaries. To separate the line boundary from the triangle boundary, the dilated boundaries will be eroded using the same structuring element of the dilation using equation (1). The algorithm will confirm if the collided object is a single boundary and also if the collided object is a line to consider the boundary as a roller support (Figure 5c). Next, the algorithm will study the slope of the line to designate the direction for the degree of freedom of the roller support.

### 2.5. The segmentation results

Figure 8 shows the overall segmentation results. Joint and line elements are shown in red, arrows are in yellow and the supports are in green. The segmentation of the pinned support from the roller is conducted and the degree of freedom for the roller support is determined as zero degree.

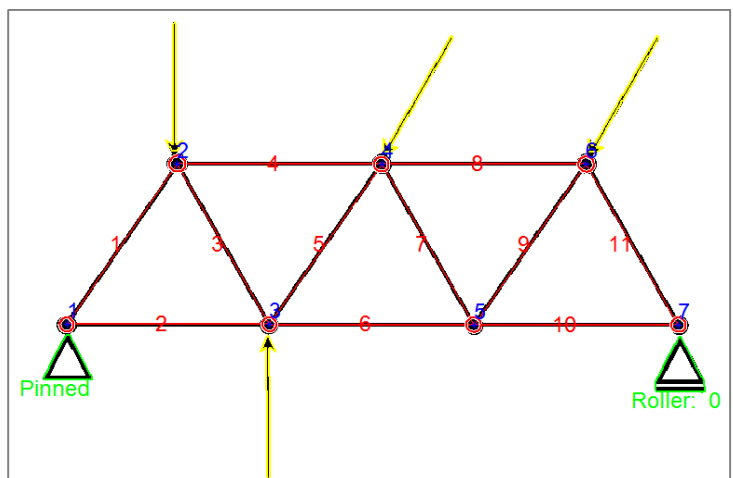

**Figure 8. The overall segmentation of the truss components**

# 3. Detection and recognition of the character strings and their assignment to the proper components

The implementation of the Optical Character Recognition (OCR) was one of the toughest challenges in this research. OCR algorithms read the letters and words from the images. The very common example of use of the OCR algorithm in the daily life is the license plate recognition to register the license plate of the heavy traffic flow automatically. In most of the approaches to the OCR problems, to study the letters of interest in any words, each letter boundary is compared with the trained boundary letters, one by one, assigning the letter of study a similarity score to each trained letter. The ID of the letter in the trained data with the most similarity score will be assigned to the boundary of the letter of study. The issue with this methodology is that if the boundary of interest is oriented, it can retrieve poor comparison results with the horizontally trained letters. In most of the OCR challenges, the words are aligned with a particular direction and are parallel to each other, making it easy for the algorithm to recognize the word without the need for calculating the orientation of the word. In this research all the used text strings in the images are oriented, and in fact, there is a handful of text strings in the image that are horizontally aligned. There are so many approaches that have been introduced to fulfill the recognition of arbitrarily oriented text strings. One of the approaches proposed by [2], and [3] is to recognize the characters individually and rotate them to match them with the trained data. This method does not run in a fast way and needs a specific training workflow [4]. In our research we used an algorithm to separate the text strings from each other, calculate the slope of the text string and then rotate it accordingly to be aligned horizontally, preparing it to be compared with the horizontal data [4]. For our OCR function to work correctly the following assumptions are made:

- All the text stings are assumed to be straight. The algorithm cannot handle curved text recognition.
- It only works for text strings containing more than 3 characters. To calculate the slope of the text strings at least 3 characters are needed.
- All the characters are assumed to be approximately the same size; otherwise, it would yield wrong results for the slope calculation.

The methodology for the used OCR is explained in the following sections:

## 3.1. Dilation of the boundaries to group the letters in each word

To group the letters in each word of interest and separate the words from each other to be processed individually, the word boundaries in the binary image is dilated, connecting the letter boundaries to each other. This division will allow to read the words separately and rotate them independently to align with the horizon. Figure 9 shows the sub-boundaries of binary image before and after dilation.

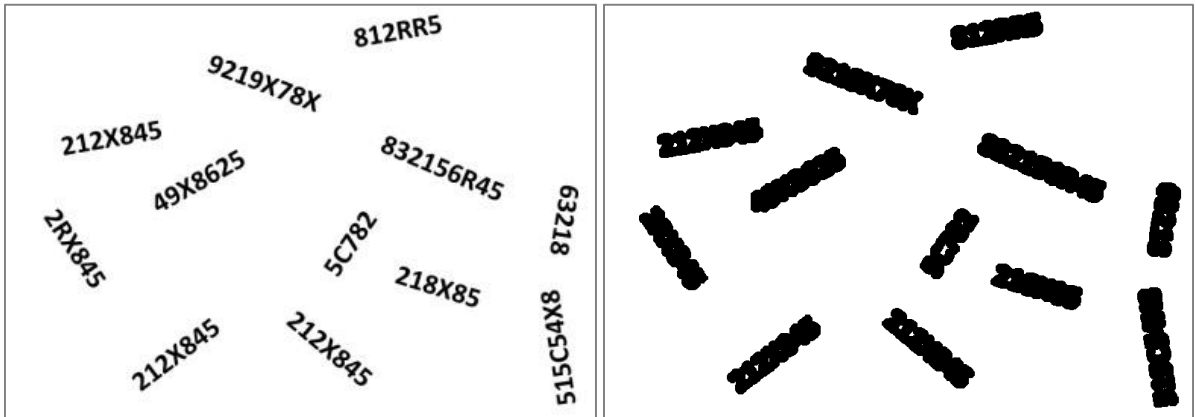

**Figure 9. The binary image (left) and the dilated image (right)**

## 3.2. Calculation of the slope for each text strings

Geometry information for all the letters in each dilated boundary is studied for their bounding-box sizes. The locations of the center of the bounding-box for each character are calculated and then are connected to each other from left to right. Then, the slope of the line connecting two neighboring bounding box centers is calculated. The overall string text slope can be calculated from the averaging the slopes between each neighbor bounding boxes letter boundaries. However, to avoid the character "." misleading the slope calculation by bringing the bounding box center down, the bounding box areas for all sub-boundaries are calculated, then, the average of the bounding box areas is considered to set a filter to eliminate the effect of the character "." on the slope calculation. In short, for a character to take part in the slope calculation of the text string, it has to have a bounding box area ranging from 0.5 to 1.4 of the average of all bounding box areas in the word [4]. Figure 10 shows the bounding boxes of a text string in blue and shows the centers of the valid boxes for the slope calculation of the text string in red.

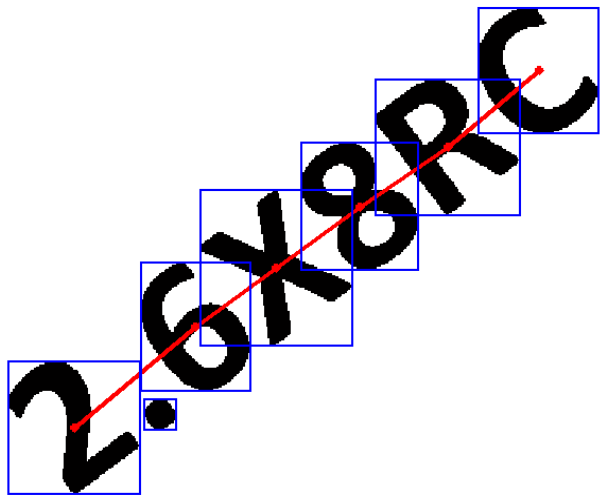

**Figure 10. Bounding boxes (in blue) and the bounding box centers (in red)**

## 3.3. The rotation of the boundary with the calculated text string slope

Each text string boundary is rotated with its associated calculated slope. This will align the text string with the horizontal.

### 3.4. The OCR function recognizing the text string

The OCR function will compare each character with trained characters to read the letter and the text string. The score result of the comparison for each character and the trained data set is recorded.

### 3.5. Checking if the text sting is not upside-down

The record of the comparison results is examined to conclude if the text string is not read upside-down. All the character correlations in each text strings are averaged. If the following criteria are met, it is concluded that the text sting is read upside-down:

- The average score result of the letter comparison in a text string is less than 0.5.
- A text string contains a character with correlation less than 0.3.

In the above condition the text string is further rotated 180 degrees and is read once more. The average score result of the comparison of the letters in 180 degrees' rotation is compared against the non-rotated one to set the correct horizontal orientation of the text string.

### 3.6. Snapping of the texts and assigning to their adjacent property

The algorithm will study the bounding box center of the dilated text strings, arrows and line boundaries to obtain their centers. It will calculate the distance of the text string centers to all object centers and will assign the content of the text string to its nearest boundary's center.

### 4. Confirmation of data validation

Any information that is missing or has been assigned wrongly to any component is spotted before sending them to Sap2000 for analysis. The algorithm will ask the user to provide the correct and proper information. The algorithm proceeds to the next steps only if all the errors are fixed by the user.

### 5. Structural analysis using Sap2000 API

An automated interaction between MATLAB and Sap2000 though API 'Application Programming Interface' has been used to utilize Sap2000's convenient analysis capability from MATLAB. Similar studies have been conducted to establish an integration between MATLAB and Sap2000 [5], [6], [7], [8]. The geometry and loading information of the truss are sent from MATLAB to Sap2000 to provide the structural analysis, and then, the analysis results are retrieved back in MATLAB. The geometry information contains the section property, support type, and location of the nodes and members. The loading information includes the load magnitude and its application orientation to the nodes. To match the data with the real scale of the problem, the user is required to provide the distance of the two nodes. All the remaining distances of the joints are scaled up and calculated accordingly. Figure 11 represents the general procedure needed to model the truss frame and conduct the structural analysis.

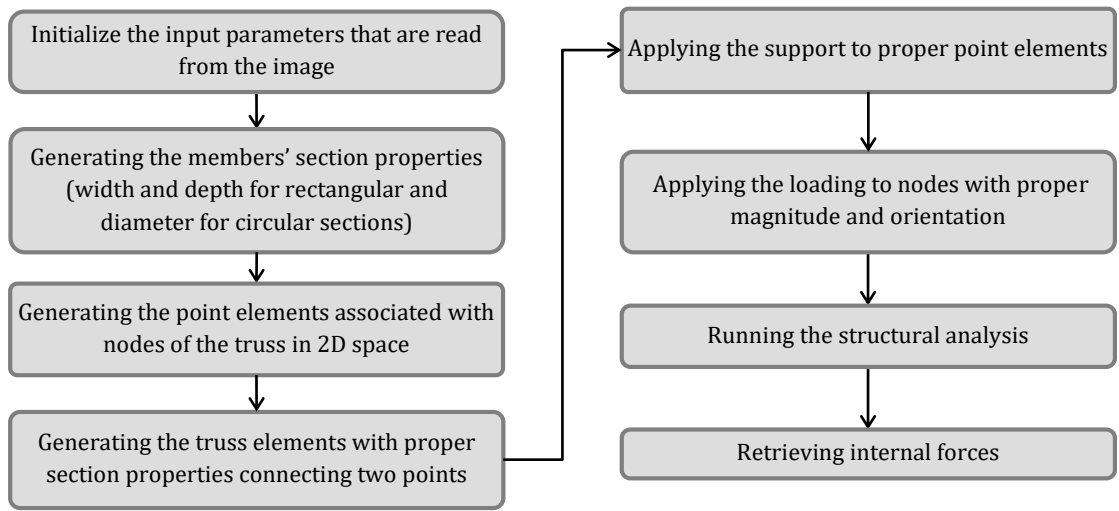
**Figure 11. The procedure to link the truss frame model to Sap2000 via API**

## 6. Overlaying the results on the input image
After conducting the structural analysis in Sap2000 the results are retrieved in MATLAB to be overlaid on the input image. Figure 12 shows all the axial loads in each element.

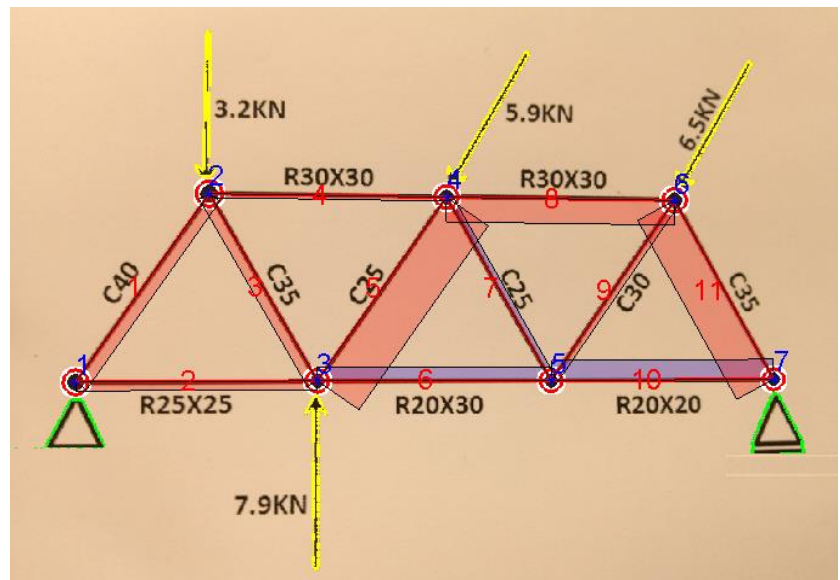
**Figure 12. The axial loads for all the members are overlaid on the input image.**

## Summery and future work
Image processing techniques were used to build a model and provide the structural analysis for a hand-sketched or a computer generated truss frame. The current work segments and analyzes the truss frame using an image as an input. This paper highlights the potential of the automated structural analysis using image processing techniques. In the future we plan to analyze the hand sketched fixed 2D rigid frames using a similar methodology. The deformed shape of a frame as well as the shear and moment forces for any member can also be visualized and overlaid on them with a similar methodology. With a further development of the current algorithm the will allow the user to quickly obtain the Building Information Model (BIM) for any available architectural drawing. Also we are interested in the methodologies of the automation in analysis of the structural frames using image processing to quickly assess and determine their seismic performance.